\documentclass[sigconf,edbt]{acmart-edbt2020}
\def\BibTeX{{\rm B\kern-.05em{\sc i\kern-.025em b}\kern-.08em
    T\kern-.1667em\lower.7ex\hbox{E}\kern-.125emX}}

\usepackage{graphicx}
\usepackage{enumitem}
\usepackage{amsmath}
\usepackage{algorithmwh}
\usepackage[T1]{fontenc}
\setcopyright{rightsretained}

\acmDOI{}

\acmISBN{978-3-89318-083-7}

\acmConference[EDBT 2020]{23rd International Conference on Extending Database Technology (EDBT)}{March 30-April 2, 2020}{Copenhagen, Denmark} 
\acmYear{2020}

\begin{document}
\title{Fairness in Clustering with \\ Multiple Sensitive Attributes}
\author{Savitha Sam Abraham$^{\bullet}$ \hspace{0.2in} Deepak P$^{\dagger, \bullet}$ \hspace{0.2in} Sowmya S Sundaram$^{\bullet}$}
\affiliation{%
  \institution{$^{\bullet}$Indian Institute of Technology Madras, India \\
  $^{\dagger}$Queen's University Belfast, UK \\
  savithas@cse.iitm.ac.in \hspace{0.2in} deepaksp@acm.org \hspace{0.2in} sowmya@cse.iitm.ac.in}
}



%
%
%


%

\begin{abstract}
A clustering may be considered as fair on pre-specified sensitive attributes if the proportions of sensitive attribute groups in each cluster reflect that in the dataset. In this paper, we consider the task of fair clustering for scenarios involving multiple multi-valued or numeric sensitive attributes. We propose a fair clustering method, \textit{FairKM} (Fair K-Means), that is inspired by the popular K-Means clustering formulation. We outline a computational notion of fairness which is used along with a cluster coherence objective, to yield the FairKM clustering method. We empirically evaluate our approach, wherein we quantify both the quality and fairness of clusters, over real-world datasets. Our experimental evaluation illustrates that the clusters generated by FairKM fare significantly better on both clustering quality and fair representation of sensitive attribute groups compared to the clusters from a state-of-the-art baseline fair clustering method. 
\end{abstract}

\maketitle              

\section{Introduction}

Clustering is the task of grouping a dataset of objects in such a way that objects that are assigned to the same group, called a {\it cluster}, are more similar to each other than those in other groups/clusters. Clustering~\cite{jain1988algorithms} is a well-studied and fundamental task, arguably the most popular task in unsupervised or exploratory data analytics. A pragmatic way of using clustering within analytics pipelines is to consider objects within the same cluster as being indistinguishable. Customers in the same cluster are often sent the same promotional material in a targeted marketing scenario in retail, whereas candidates clustered using their resumes may be assigned the same shortlisting decision in a hiring scenario. Clustering provides a natural way to tackle the infeasibility of doing manual per-object assessment or appreciation, especially in cases where the dataset in question encompasses more than a few hundreds of objects. Central to clustering is the notion of similarity which may need to be defined in a task-oriented manner. As an example, clustering to aid a task on identifying tax defaulters may use a similarity measure that focuses more on the job and income related attributes, whereas that for a health monitoring task may more appropriately focus on a very different set of attributes. 

Usage of clustering algorithms in analytics pipelines for making important decisions open up possibilities of unfairness. Among two high-level streams of fairness constructs, viz., individual fairness~\cite{dwork2012fairness} and group fairness~\cite{kamiran2009classifying}, we focus on the latter. Group fairness considers fairness from the perspective of {\it sensitive attributes} such as gender and ethnicity and groups defined on the basis of such sensitive attributes. Consider a clustering algorithm that targets to group records of individuals to clusters; it is possible and likely that certain clusters have highly skewed distributions when profiled against particular sensitive attributes. As an example, clustering a dataset with broad representation across gender groups based on exam scores could lead to clusters that are highly gender-homogeneous due to statistical correlations\footnote{https://www.compassprep.com/new-sat-has-disadvantaged-female-testers/}; this would happen even if the gender attribute were not explicitly considered within the clustering, since such information could be held implicitly across one or more other attributes. Choosing a cluster with a high gender or ethnic skew for positive (e.g., interview shortlisting) or negative (e.g., higher scrutiny or checks) action entails differentiated impact across gender and ethnic groups. This could also lead to reinforcement of societal stereotypes. For example, consistently choosing individuals from particular ethnic groups for pro-active surveillance could lead to higher reporting of violations from such ethnic groups since enhanced surveillance translates to higher crime visibility, resulting in higher reporting rates. This reporting skew results thus manifests as a data skew which provides opportunities for future invocations of the same analytics to exhibit higher ethnic asymmetry. In modern decision making scenarios within a liberal and democratic society, we need to account for a plurality of sensitive attributes within analytics pipelines to avoid causing (unintended) discrimination; these could include gender, race, religion, relationship status and country of origin in generic decision-making scenarios, and could potentially include attributes such as age and income in more specific ones. There has been much recent interest in the topic of fair clustering~\cite{chierichetti2017fair,ahmadian2019clustering}. 

\subsubsection*{Our Contribution}

In this paper, we consider the task of fair clustering in the presence of multiple sensitive attributes. As will be outlined in a later section, this is a direction that has received less attention amidst existing work on fair clustering that has been designed for scenarios involving a single binary sensitive attribute~\cite{chierichetti2017fair,kleindessner2019guarantees,backurs2019scalable,olfat2019convex}, single multi-valued/categorical sensitive attribute~\cite{ahmadian2019clustering,wang2019towards,ziko2019clustering,schmidt2018fair}, or multiple overlapping groups~\cite{bera2019fair}. We propose a clustering formulation and algorithm to embed group fairness in clustering that incorporates multiple sensitive attributes that may include numeric, binary or multi-valued (i.e., categorical) sensitive ones. Through an empirical evaluation over multiple datasets, we illustrate the empirical effectiveness of our approach in generating clusters with fairer representations of sensitive groups, while preserving cluster meaningfulness.


\section{Related Work}
Fairness in machine learning has received significant attention over the last few years. Our work contributes to the area of fair methods for unsupervised learning tasks. We now briefly survey the area of fairness in unsupervised learning, with a focus on clustering, our task of interest. We interchangeably refer to groups defined on sensitive attributes (e.g., {\it ethnic groups}, {\it gender groups} etc.) as {\it protected classes} for consistency with terminology in some referenced papers. Towards developing a systematic summary, we categorize prior work into three types depending on whether the fairness modelling appears as a (i) pre-processing step, (ii) during the task of clustering, or (iii) as a post-processing step after clustering. These yield the three technique families.

\begin{table*}[]
\centering
\begin{tabular}{p{1.3 cm}llp{11.7 cm}}
\\
\cline{1-4}
{\bf Paper} &{\bf Number} & {\bf Type} & {\bf Fairness Definition} \\
\cline{1-4}
\cite{chierichetti2017fair},\cite{backurs2019scalable},\cite{anagnostopoulos2019principal}  &  Single & Binary & Preserve proportional representation of protected classes within clusters. \\ 
\cite{schmidt2018fair} & Single & Multi-valued & Preserve proportional representation of protected classes within clusters. \\ 
 \cite{olfat2019convex} & Single & Binary & The accuracy of the classifier predicting the protected class of a data point should be within a specified bound.\\
\cite{wang2019towards} & Single & Multi-valued & Each cluster should have an equal number of data points from each protected class.\\
\cite{bera2019fair} & Multiple& Binary & The proportional representation of a protected class in a cluster should be within the specified lower and upper bounds.\\
\cite{ahmadian2019clustering}&Single &Multi-valued&The proportional representation of a protected class in a cluster should not go beyond a specified upper bound.\\
\cite{kleindessner2019fair}&Single&Multi-valued&The clustering should produce pre-specified number of cluster centers belonging to each specific protected class.\\
\cite{kleindessner2019guarantees}, \cite{ziko2019clustering}&Single&Multi-valued& Preserve proportional representation of protected classes within clusters. \\ 
\cite{chen2019proportionally}&-&-& There are no set of $(n/k)$ points such  that there exists another center that is closer to each of these $(n/k)$ points.\\
\cite{rosner2018privacy}&Single&Multi-valued& Each cluster should have atleast a pre-specified number of points of a protected class. \\
\textbf{FairKM}&Multiple&Multi-valued/&Preserve proportional representation of protected classes within clusters. \\
&&Numeric&as its representation in the whole dataset.\\
\cline{1-4}
\end{tabular}
\caption{Fair Unsupervised ML Methods indicating the Number and Type of Sensitive Attributes they are designed for.}
\label{tab:rel}
\end{table*}


\subsection{Space Transformation Approaches} 

The family of fairness pre-processing techniques work by first representing the data points in a {\it `fair'} space followed by applying any existing clustering algorithm upon them. This is the largest family of techniques, and most approaches in this family seek to achieve theoretical guarantees on representational fairness. This family includes one of the first works in this area of fair clustering~\cite{chierichetti2017fair}. The work proposes a fair variant of classical clustering for scenarios involving a single sensitive attribute that can take binary values. Let each object be colored either {\it x} or {\it y} depending on its value for the binary sensitive attribute. \cite{chierichetti2017fair} defines fairness in terms of \textit{balance} of a cluster, which is $min\{\#x/\#y,\#y/\#x\}$. They go on and outline the concept of $(b,r)$-\textit{fairlet decomposition}, where the points in the dataset are grouped into small clusters called fairlets, such that each fairlet would have a balance of $b/r$, where $b<r$. The clustering is then performed on these fairlets. The fairness guarantees that are provided by the clustering is based on the balance in the underlying fairlets. Fairlet decomposition turns out to be NP-hard, for which an approximation algorithm is provided. Later, \cite{backurs2019scalable} proposed a faster fairlet decomposition algorithm offering significant speedups. The work in \cite{schmidt2018fair} extends the fairlet idea to $K$-means for scenarios with a single multi-valued sensitive attribute. They define fair-coresets which are smaller subsets of the original dataset, such that solving fair clustering over this subset also results in giving an approximate solution for the entire dataset.

Another set of fair space transformation approaches build upon methods for dimensionality reduction and representation learning. A recent such work \cite{anagnostopoulos2019principal} considers the bias in the dataset (that is, representational skew) as a form of noise and uses spectral de-noising techniques to project the original points in the dataset to a new fair projected space. \cite{olfat2019convex} describes a fair version of PCA (Principal Component Analysis) for data with a single binary sensitive attribute. A representation learning method may be defined as fair if the information about the sensitive attributes cannot be inferred from the learnt representations. The method uses convex optimization formulations to embed fairness within PCA. The fairness is specified in terms of failure of the classifiers in predicting the sensitive class of the dimensionality-reduced data points obtained from fair PCA. The method is fair{\it er} if the data points are less distinguishable with respect to their values of the sensitive attribute in this lower dimension space. Another work that projects the original data points into a fair space is the one described in \cite{wang2019towards}. This method, which is for cases involving a single multi-valued sensitive attribute, defines a clustering to be fair when there is an equal number of data points of each protected class in each cluster. They define the concept of \textit{fairoids} (short for {\it fair centroids}) which are formed by grouping together all points belonging to the same protected class. The task is then to learn a latent representation of the data points such that the cluster centroids obtained after clustering on this latent representation are equi-distant from every fairoid. 


\subsection{Fairness Modelling within Optimization}

Methods in this family incorporate the fairness constraints into the clustering step, most usually within the objective function that the clustering method seeks to optimize for. It may be noted that the method we propose in this paper, {\it FairKM}, also belongs to this family. Approaches within this family define a clustering to be fair if the proportional representation of the protected class in a cluster reflects that in the dataset. One of the techniques~\cite{kleindessner2019guarantees} describes a fair variant of spectral clustering where a linear fairness constraint is incorporated into the original ratio-cut minimization objective of spectral clustering. Another recent technique~\cite{ziko2019clustering}, the method that comes closest to ours in construction, modifies $K$-means clustering to add a fairness loss term. The fairness loss is computed as the KL-divergence between the probability distribution across the different values for the sensitive attribute in a cluster, and the corresponding distribution for the whole dataset. This method is designed for a single multi-valued sensitive attribute and does not generalize to multiple such sensitive attributes. {\it Being closest to our proposed method in spirit, we use this method as our primary baseline, in the experiments.} 

In contrast to the above, another recent work~\cite{chen2019proportionally} outlines a different notion of fairness, one that is independent of (and agnostic to) sensitive attributes. They define fairness as proportionality wrt spatial distributions, to mean that any $(n/k)$ points can form their own cluster if there exists another center that is closer to each of these $(n/k)$ points. This proportionality constraint is incorporated into the objective function of k-median clustering and is optimized to find a clustering that satisfies this constraint. 

\subsection{Cluster Perturbation Techniques}

In this third family of techniques, vanilla clustering is first applied on the original set of data points, after which the generated clusters are perturbed to improve fairness of the solution. In \cite{bera2019fair}, fairness is defined in terms of a lower and upper bound on the representation of a protected class in a cluster. This method is for cases with multiple binary sensitive attributes, referred to as {\it overlapping groups} in the paper. The k-centers generated from vanilla clustering on the data points are used to perform a fair partial assignment between points and the centers. The fair partial assignment is formulated as a linear program with constraints that ensures that the sum of the weights associated with a point's partial assignments is one, and, the representation of a protected class in a cluster is within the specified upper and lower bounds. The partial assignments are then converted to integral assignments by framing it as another linear programming problem. \cite{ahmadian2019clustering} also uses a similar idea, 
but it just enforces an upper bound, consequently preventing over-representation of specific groups in a cluster. The work described in \cite{kleindessner2019fair} proposes a simple approximation algorithm for k-center clustering under a fairness constraint, for scenarios with a single multi-valued sensitive attribute. The method targets to generate a fair summary of a large set of data points, such that the summary is a representative subset of the original dataset. For example, if the original dataset has a 70:30 male:female distribution, then a fair summary should also have the same distribution. 


\subsection{Summary}

Table \ref{tab:rel} summarizes the different approaches in literature and our proposed approach {\it FairKM}, in terms of the number and type of sensitive attributes they handle and their definition for fairness. As it may be seen from the table, there has been very limited exploration into methods that admit multiple multi-valued ({\it aka} categorical or multi-state) sensitive attributes, the space that {\it FairKM} falls in. While multiple multi-valued attributes can be treated as together forming a giant multi-valued attribute taking values that are combinations of the component attributes, this results in a large number of very fine-grained groupings. These make it infeasible to both (i) impose fairness constraints over, and (ii) ensure parity in treatment of different sensitive attributes independent of the differences in the number of values they take. Considering the fact that real-life scenarios routinely present with multiple sensitive attributes, {\it FairKM}, we believe addresses an important line of inquiry in the backdrop of the literature. We will empirically evaluate {\it FairKM} against~\cite{ziko2019clustering}, the latter coming from the same technique family and having similar construction. 


\section{Problem Definition}\label{sec:probdef}

Let $\mathcal{X} = \{ \ldots, X, \ldots \}$ be a dataset of records defined over two sets of attributes $\mathcal{N}$ and $\mathcal{S}$. $\mathcal{N}$ stands for the set of attributes that are relevant to the task of interest, and thus may be regarded non-sensitive. As examples, this may comprise experience and skills in the case of screening applicants for a job application, and attributes from medical wearables' data to inform decision making for pro-active screening. $\mathcal{S}$ stands for the set of {\it sensitive attributes}, which would typically include attributes such as those identifying {\it gender}, {\it race}, {\it religion}, {\it relationship status} in a citizen database and any other sensitive attribute over which fairness is to be ensured. In other scenarios such as NLP for education, representational fairness may be sought over attributes such as types of problems in a word problem database; this forms one of the scenarios in our empirical evaluation. 

The (vanilla) clustering objective would be to group $\mathcal{X}$ into clusters such that it maximizes intra-cluster similarity and minimizes inter-cluster similarity, similarity gauged using the task-relevant attributes in $\mathcal{N}$. Within a fair clustering, we would additionally like the output clusters to be {\it fair} on attributes in $\mathcal{S}$. A natural way to operationalize fairness within a clustering that covers all objects in $\mathcal{X}$ would be to ensure that the distribution of groups defined on sensitive attributes within each cluster approximates the distribution across the dataset; this correlates with the well-explored notion of {\it statistical parity}~\cite{dwork2012fairness} in fair supervised learning. For example, suppose the sex ratio in $\mathcal{X}$ is 1:1; we would ideally like each cluster in the clustering output to report a sex ratio of 1:1, or very close to it. In short, we would like a fair clustering to produce clusters, each of which are both:

\begin{itemize}[leftmargin=*]
\item coherent when measured on the attributes in $\mathcal{N}$, {\it and}
\item approximate the dataset distribution when measured on the attributes in $\mathcal{S}$.
\end{itemize}

It may be noted that simply {\it hiding} the $\mathcal{S}$ attributes from the clustering algorithm does not suffice. A {\it gender blind} clustering algorithm may still produce clusters that are highly gender-homogeneous, since some attributes in $\mathcal{N}$ could implicitly encode gender information. Indeed, we would like a fair clustering to surpass $\mathcal{S}$-blind clustering by significant margins on fairness.


\section{FairKM: Our Method}

We now describe our proposed technique for fair clustering, code-named {\it FairKM}, short for Fair K-Means, indicating that it draws inspiration from the classical $K$-Means clustering algorithm~\cite{macqueen1967some,jain2010data}. FairKM incorporates a novel fairness loss term that nudges the clustering towards fairness on attributes in $\mathcal{S}$. The FairKM objective function is as follows:

\begin{multline}
\mathcal{O} = \\
\underbrace{\sum_{C \in \mathcal{C}} \sum_{X \in C} dist_{\mathcal{N}}(X,C)}_\text{K-Means Term over attributes in $\mathcal{N}$} + \lambda \underbrace{ deviation_{\mathcal{S}}(\mathcal{C},\mathcal{X})}_\text{Fairness Term over attributes in $\mathcal{S}$}
\label{eq:objective}
\end{multline}


As indicated, the objective function comprises two components; the first is the usual $K$-Means loss for the clustering $\mathcal{C}$, $dist_{\mathcal{N}}(X,C)$ computing the distance between $X$ and prototype of cluster $C$, distance computed only over attributes in $\mathcal{N}$. The second is a fairness loss term we introduce, which is computed over attributes in $\mathcal{S}$. $\lambda$ is a parameter that may be used to balance the relative strength of the two terms. As in $K$-Means, this loss is computed over a given clustering; the task is thus to identify a clustering that minimizes $\mathcal{O}$ as much as possible. We now describe the details of the second term, and the intuitions behind its construction. 

\subsection{The Fairness Term in FairKM}

While the $K$-Means term in the {\it FairKM} objective tries to ensure that the output clusters are coherent in the $\mathcal{N}$ attributes, the fairness term performs the complementary function of ensuring that the clusters manifest fair distributions of groups defined by sensitive attributes in $\mathcal{S}$. We outline the motivation and construction of the fairness term herein. 

\noindent{\bf Attribute-Specific Deviation for a Cluster:} Consider a single sensitive attribute $S$ (e.g., gender) among the set of sensitive attributes $\mathcal{S}$. For each data object $X$, $S$ may take on one value from a set of permissible values. Let $s$ be one such value (e.g., female, for the choice of $S$ as gender). For an ideally fair cluster $C$, one would expect that the fractional representation of $s$ in $C$ - the fraction of objects in $C$ that take the value $s$ for $S$ - to be close to the fractional representation of $s$ in $\mathcal{X}$. With our intent of generating clusters that are as fair as possible, we seek to generate clusterings such that the deviation between the fractional representations of $s$ in $C$ and $\mathcal{X}$ are minimized for each cluster $C$. For a given cluster $C$ and a choice of value $s$ for $S$, we model the deviation as simply the square of the absolute differences between the fractional representations in $C$ and $\mathcal{X}$:

\begin{equation}
    D_C^S(s) = \bigg( \frac{|\{X| X \in C \wedge X.S = s\}|}{|C|} - \frac{|\{X| X \in \mathcal{X} \wedge X.S = s\}|}{|\mathcal{X}|} \bigg)^2
\end{equation}

The deviation, when aggregated over all values of $S$, yields:

\begin{equation}
    D_C^S = \begin{cases}
    \sum_{s \in Values(S)} D_C^S(s) & |C| \neq 0 \\
    0 & |C| = 0
    \end{cases}
\end{equation}

The above aggregation accounts for the fact that $D_C^S(s)$ is undefined when $C$ is an empty cluster. 

\noindent{\bf Domain Cardinality Normalization:} Different sensitive attributes may have different numbers of permissible values (or domain cardinalities). For example, {\it race} and {\it gender} attributes typically take on much fewer values than {\it country of origin}. Those attributes with larger domains are likely to yield larger $D_C^S$ scores because, (i) the deviations are harder to control within (small) clusters given the higher likely scatter, and (ii) there are larger numbers of $D_C^S(s)$ terms that add up to $D_C^S$. In order to ensure that attributes with larger domains do not dominate the fairness term, we normalize the deviation by the number of different values taken by an attribute, yielding $ND_C^S$, a normalized attribute-specific deviation:

\begin{equation}
    ND_C^S = \frac{D_C^S}{|Values(S)|}
\end{equation}

This is then summed up over all attributes in $\mathcal{S}$ to yield a single term for each cluster:

\begin{equation}
    ND_C = \sum_{S \in \mathcal{S}} ND_C^S
\end{equation}

\noindent{\bf Cluster Weighting:} Observe that $ND_C$ deviation loss would tend towards $0.0$ for very large clusters, since they are obviously likely to reflect dataset-level distributions better; further, an empty cluster would also have $ND_C^S = 0$ by definition. Considering the above, a clustering loss term modelled as a simple sum over its clusters, $\big( \sum_{C \in \mathcal{C}} ND_C \big)$ or a cardinality weighted sum, $\big( \sum_{C \in \mathcal{C}} |C| \times ND_C \big)$, can both be driven towards $0.0$ by keeping a lot of clusters empty, and distributing the dataset objects across very few clusters; the boundary case would be a single non-empty cluster. Indeed, this propensity towards the boundary condition is kept in check by the $K$-Means term; however, we would like our fairness term to drive the search towards more reasonable fair clustering configurations in lieu of simply reflecting a propensity towards highly skewed clustering configurations. 

Towards achieving this, we weight each cluster's deviation by the square of it's fractional cardinality of the dataset. This leads to an overall loss term as follows:

\begin{equation}
    \sum_{C \in \mathcal{C}} \bigg( \frac{|C|}{|\mathcal{X}|} \bigg)^2 \times ND_C
\end{equation}

The squared term in the weighting enlarges the $ND_C$ terms of larger clusters much more than smaller ones, making it unprofitable to create large clusters; this compensates for the propensity towards skewed clusters as embodied in the loss construction. 

\noindent{\bf Overall Loss:} The overall fairness loss term is thus:

\begin{multline}
    deviation_{\mathcal{S}}(\mathcal{C},\mathcal{X}) = \\
    \sum_{C \in \mathcal{C}} \bigg( \frac{|C|}{|\mathcal{X}|} \bigg)^2 \times \sum_{S \in \mathcal{S}} \frac{\sum_{s \in Values(S)} \big( Fr_C^S(s) - Fr_{\mathcal{X}}^S(s) \big)^2}{|Values(S)|}
    \label{eq:fairnessterm}
\end{multline}

where $Fr_C^S(s)$ and $Fr_{\mathcal{X}}^S(s)$ are shorthands for the fractional representation of $S=s$ objects in $C$ and $\mathcal{X}$ respectively.

\subsection{The Optimization Approach}

Having defined the objective function, we now outline the optimization approach. It is easy to observe that there are three sets of parameters, the clustering assignments for each data object in $\mathcal{X}$, the cluster prototypes that are used in the first term of the objective function, and the fractional representations, i.e., $Fr_{C}^S(s)$s, used in the fairness term. Unlike $K$-Means, given the more complex construction, it is harder to form a closed-form solution for the cluster assignments. Thus, from a given estimate of all three sets of parameters, we step over each data object $X \in \mathcal{X}$ in round-robin fashion, updating its cluster assignment, and making consequent changes in cluster prototypes and fractional representations. One set of round-robin updates forms an iteration, with multiple such iterations performed until convergence or until a maximum threshold of iterations is reached. 








\subsubsection{Cluster Assignment Updates}

At $\lambda = 0$, {\it FairKM} defaults to $K$-Means where the cluster assignments are determined only by proximity to the cluster prototype (over attributes in $\mathcal{N}$). At higher values of $\lambda$, {\it FairKM} cluster assignments are increasingly swayed by considerations of representational fairness of $\mathcal{S}$ attributes within clusters. 

It may be noted that the cluster assignments are used in both the terms of the {\it FairKM} objective, in different ways. This makes a closed form estimation of cluster assignments harder to arrive at. This leads us to a round-robin approach of determining cluster assignments. When each $X$ is considered, the cluster prototypes as well as the current cluster assignments of all other objects, i.e. $\mathcal{X} - \{X\}$, are kept unchanged. The cluster assignment for $X$ is then estimated as:


\begin{equation}
    Cluster(X) = \mathop{\arg\min}_{C} \ \ \mathcal{O}_{\mathcal{C} + (X \in C)}
    \label{eq:estep}
\end{equation}

For the candidate object $X$, we evaluate the value of the objective function by changing $X$'s cluster membership from the present one to each cluster, $\mathcal{C} + (X \in C)$ indicating a corresponding change in the clustering configuration retaining all other objects' present cluster assignments. $X$ is then assigned to the cluster for which the minimum value of $\mathcal{O}$ is achieved. While this may look as a simple step, implementing it naively is computationally expensive. However, easy optimizations are possible when one observes the dynamics of the change and how it operates across the two terms. We now outline a simple way of affecting the cluster assignment decision. First, let $X$'s current cluster assignment be $C'$; the cluster assignment step can be equivalently written as:

\begin{equation}
    Cluster(X) = \mathop{\arg\min}_{C} \delta \mathcal{(O)}_{X \in C' \rightarrow X \in C}
    \label{eq:estepoptim}
\end{equation}

where $\delta\mathcal{O}$ indicates the change in $\mathcal{O}$ when the respective cluster assignment change is carried out. This can be expanded into the changes in the two terms in the objective function as follows:

\begin{multline}
    \delta \mathcal{(O)}_{X \in C' \rightarrow X \in C} = \\ \delta \text{(K-Means term)}_{X \in C' \rightarrow X \in C} + \lambda \times \delta \text{(deviation term)}_{X \in C' \rightarrow X \in C}
    \label{eq:estep1}
\end{multline}

We now detail the changes in the respective terms separately. 

\noindent{\bf Change in K-Means Term:} We now outline the change in the $K$-Means term by moving $X$ from $C'$ to $C$. As may be obvious, this depends only on attributes in $\mathcal{N}$. We model the cluster prototypes as simply the average of the objects within the cluster. The change in the $K$-Means term is the sum of (i) the change in the $K$-Means term corresponding to $C'$ brought about by the exclusion of $X$ from it, and (ii) the change in the $K$-Means term corresponding to $C$ brought about by the inclusion of $X$ within it. We discuss them below. 

Consider a single numeric attribute $N \in \mathcal{N}$, for simplicity. Through excluding $X$ from $C'$, the $N$ attribute value of the cluster prototype of $C'$ undergoes the following change:

\begin{equation}
    C'.N \rightarrow \bigg[ \bigg(C'.N - \frac{X.N}{|C'|}\bigg) \times \frac{|C'|}{|C'-1|} \bigg]
    \label{eq:newcprimen}
\end{equation}

where $C'$ is overloaded to refer to the cluster and the cluster prototype (to avoid clutter), all values referring to those prior to exclusion of $X$. The term after the $\rightarrow$ stands for the $N$ attribute value for the new cluster prototype. As indicated, it is computed by the removal of the contribution from $X$ from the cluster prototype, followed by re-normalization, now that $C'$ has one fewer object within it. The change in the $K$-Means term for $N$ corresponding to $C'$ is then as follows:

\begin{multline}
    \delta_{X_{out}}KM(C',N) = \bigg( \sum_{X' \in C', X' \neq X} (X'.N - New(C'.N))^2 \bigg) - \\ \bigg[ \bigg( \sum_{X' \in C', X' \neq X} (X'.N - C'.N)^2 \bigg) + (X.N - C'.N)^2 \bigg]
    \label{eq:kmout}
\end{multline}

where $New(C'.N)$ is the new estimate of $C'.N$ as outlined in Eq.~\ref{eq:newcprimen}. The first term corresponds to the $K$-Means loss in the new configuration (after exclusion of $X$), whereas the sum of the second and third terms correspond to that prior to exclusion of $X$. Analogous to the above, the new centroid computation for $C$ and the change in the $K$-Means terms are outlined below:

\begin{equation}
    C.N \rightarrow \bigg[ \bigg( C.N \times \frac{|C|}{|C+1|} \bigg) + \frac{X.N}{C+1} \bigg]
\end{equation}

\begin{multline}
    \delta_{X_{in}}KM(C,N) = \bigg[ \bigg( \sum_{X' \in C, X' \neq X} (X'.N - New(C.N))^2 \bigg) + \\ (X.N - New(C.N))^2 \bigg] - \bigg( \sum_{X' \in C, X' \neq X} (X'.N - C.N)^2 \bigg) \\ 
    \label{eq:kmin}
\end{multline}

It may be noticed that the computation of the changes above only involve $X$ and other objects in $C$ and $C'$. In particular, the other clusters and their objects do not come into play. So far, we have computed the changes for only one attribute $N$. The overall change in the $K$-Means term is simply the sum of these changes across all attributes in $\mathcal{N}$. 

\begin{multline}
    \delta \text{(K-Means term)}_{X \in C' \rightarrow X \in C} = \\ \sum_{N \in \mathcal{N}} \bigg( \delta_{X_{out}}KM(C',N) + \delta_{X_{in}}KM(C,N) \bigg)
\end{multline}




\noindent{\bf Change in Fairness Term:} We now outline the construction of the change in the fairness term. As earlier, we start by considering a single cluster $C^*$, a single attribute $S$, and a single value $s$ within it. The fairness term from Eq.~\ref{eq:fairnessterm} can be written as follows:

\begin{equation}
    dev(C^*,S = s) = \frac{C^{*^2} \times \bigg( \bigg( \frac{C^*_s}{C^*} \bigg)^2 + \bigg( \frac{\mathcal{X}_s}{\mathcal{X}} \bigg)^2 -2 \frac{C^*_s \ \mathcal{X}_s}{C^* \ \mathcal{X}}\bigg)}{\mathcal{X}^2 \times |Values(S)|}
\end{equation}

where each set ($C^*$ and $\mathcal{X}$) is overloaded to represent both itself and its cardinality (to avoid notation clutter), and their suffixed versions ($C^*_s$ and $\mathcal{X}_s$) are used to refer to their subsets containing their objects which take the value $S = s$. The above equation follows from the observation that $Fr_{C^*}^S(s) = \frac{C^*_S}{C^*}$ and analogously for $\mathcal{X}$. When an object changes clusters from $C'$ to $C$, there is a change in the terms associated with both clusters, as in the previous case. The change in the origin cluster $C'$ works out to be the follows:

\begin{multline}
    \delta_{X_{out}} dev(C',S = s) = \frac{1}{\mathcal{X}^2 \times |Values(S)|} \times 
    \bigg[ \bigg(\frac{\mathcal{X}_s}{\mathcal{X}}\bigg)^2 (1 - 2 C') + \\ \mathbb{I}(X.S = s)(1 - 2 C'_s) - 2 \bigg( \frac{\mathcal{X}_s}{\mathcal{X}} \bigg) \bigg(\mathbb{I}(X.S = s)(1-C')-C'_s\bigg) \bigg]
    \label{eq:fairnessout}
\end{multline}

where $\mathbb{I}(.)$ is an indicator function, and $C'$ and $C'_s$ denote the cardinalities before $X$ is taken out of $C'$. We omit the derivation for space constraints. Intuitively, to nudge clusters towards fairness, we would like to incentivize removal of objects with $S=s$ from $C'$ when $C'$ is overpopulated with such objects (i.e., $C'_s$ is high). This is evident in the $-(C'_s \times \mathbb{I}(X.S=s))$ component; when $C'_s$ is high, removal of an object with $S=s$ entails a bigger reduction in the objective. The analogous change in the target cluster $C$, is as follows:

\begin{multline}
    \delta_{X_{in}} dev(C,S=s) = \frac{1}{\mathcal{X}^2 \times |Values(S)|} \times \bigg[ \bigg(\frac{\mathcal{X}_s}{\mathcal{X}}\bigg)^2 (1 + 2 C) + \\
    \mathbb{I}(X.S = s)(1 + 2 C_s) - 2 \bigg( \frac{\mathcal{X}_s}{\mathcal{X}} \bigg) \bigg(\mathbb{I}(X.S = s)(1+C)+C_s\bigg) \bigg]
    \label{eq:fairnessin}
\end{multline}

where $C$ and $C_s$ denote the cardinalities before $X$ is inserted into $C$. Given that we are inserting $X$ into $C$, the fairness intuition suggests that we should disincentivize addition of objects with $s$ when $C$ already has too many of such objects. This is reflected in the $(C_s \times \mathbb{I}(X.S=s))$ term; notice that this is exactly the same term as in the earlier case, but with a different sign. 

Thus, the overall fairness term change is as follows:

\begin{multline}
    \delta \text{(deviation term)}_{X \in C' \rightarrow X \in C} = \\ \sum_{S \in \mathcal{S}} \sum_{s \in Values(S)} \bigg( \delta_{X_{out}} dev(C',S = s) + \delta_{X_{in}} dev(C,S=s) \bigg)
    \label{eq:estep10}
\end{multline}

This completes all the steps required for Eq.~\ref{eq:estepoptim}. Based on the change in the cluster assignment, the cluster prototypes and fractional representations are to be updated. 

\subsubsection{Cluster Prototype Updates}\label{sec:clusterprototypeupdates}

Once a new cluster has been finalized for $X$, the origin and target cluster prototypes are updated according to Eq.~\ref{eq:kmout} and Eq.~\ref{eq:kmin} respectively. 

\subsubsection{Fractional Representation Updates}

The $Fr_{C'}^S(s)$s and $Fr_{C}^S(s)$s need to be updated to reflect the change in the cluster assignment of $X$. These are straightforward and given as follows:

\begin{equation}
    \forall S \forall s \in Values(S), Fr_{C'}^S(s) = 
    \begin{cases}
        \frac{C'_s - 1}{C' - 1} & \text{if } X.S = s \\
        \frac{C'_s}{C' - 1} & \text{if } X.S \neq s \\
    \end{cases}
    \label{eq:fractionold}
\end{equation}

\begin{equation}
    \forall S \forall s \in Values(S), Fr_{C}^S(s) = 
    \begin{cases}
        \frac{C_s + 1}{C + 1} & \text{if } X.S = s \\
        \frac{C_s}{C' + 1} & \text{if } X.S \neq s \\
    \end{cases}
    \label{eq:fractionnew}
\end{equation}

where the $C$, $C'$, $C_s$ and $C'_s$ values above are cardinalities of the respective sets prior to the update to $X$'s cluster assignment. 

\begin{algorithm}
\caption{\bf \textit{FairKM}}
\begin{flushleft}
Input. Dataset $\mathcal{X}$, attribute sets $\mathcal{S}$ and $\mathcal{N}$, number of clusters $k$ \\
Hyper-parameters: Fairness Weighting $\lambda$ \\
Output. Clustering $\mathcal{C}$
\end{flushleft}
\begin{code}
1. {\it Initialize $k$ clusters randomly} \\
2. {\it Set cluster prototypes as Cluster Centroids} \\
3. $while(not\ yet\ converged\ and\ max.\ iterations\ not\ reached)$ \\
4. \> $\forall X \in \mathcal{X}$, \\
5. \> \> {\it Set Cluster(X) using Eq.~\ref{eq:estepoptim}} (and Eq.~\ref{eq:estep1} through Eq.~\ref{eq:estep10}) \\
6. \> \> {\it Update cluster prototypes as outlined in Sec~\ref{sec:clusterprototypeupdates}} \\
7. \> \> {\it Re-estimate the $Fr_C^S(s)$ using Eq.~\ref{eq:fractionold} and Eq.~\ref{eq:fractionnew}} \\
8. {\it Return the current clustering assignments as }$\mathcal{C}$
\end{code}
\label{alg:fairkm}
\end{algorithm}

\subsection{FairKM Algorithm}

Having outlined the various steps, the {\it FairKM} algorithm can now be summarized in Algorithm~\ref{alg:fairkm}. The method starts with a random initialization of clusterings (Step 1) and proceeds iteratively. Within each iteration, each object is considered in round-robin fashion, executing three steps in sequence: (i) updating the cluster assignment of $X$ (Step 5), (ii) updating the cluster prototypes to reflect the change in cluster assignment of $X$ (Step 6), and (iii) updating the fractional representations correspondingly (Step 7). The significant difference in construction from $K$-Means is due to the inter-dependency in cluster assignments; the cluster assignment for $X$ depends on the current cluster assignments for all other objects $\mathcal{X} - \{X\}$, due to the construction of the {\it FairKM} objective as reflected in the update steps. The updates proceed as long as the clustering assignments have not converged or a pre-specified maximum number of iterations have not reached. 

\subsubsection{Complexity:} The time complexity of {\it FairKM} is dominated by the cluster assignment updates. Within each iteration, for each $X$ ($|\mathcal{X}|$ of them) and each cluster it could be re-assigned to ($k$ of them), the deviation needs to be computed for both the (i) $K$-Means term, and the (ii) fairness term. First, considering the $K$-Means term, it may be noted that each other object in $\mathcal{X}$ would come into play once, either as a member of $X$'s current cluster (in Eq.~\ref{eq:kmout}) or as a member of a potential cluster to which $X$ may be assigned (in Eq.~\ref{eq:kmin}). This yields an overall complexity of each $K$-Means deviation computation being in $\mathcal{O}(|\mathcal{X}||\mathcal{N}|)$. Second, considering the fairness deviation computation, it may be seen as a simple computation (Eq.~\ref{eq:fairnessout} and~\ref{eq:fairnessin}) that can be completed in constant time. This computation needs to be performed for each attribute in $\mathcal{S}$ and each value of the attribute (consider $m$ as the maximum number of values across attributes in $\mathcal{S}$), yielding a total complexity of $\mathcal{O}(|\mathcal{S}|m)$ for each fairness update computation. With the updates needing to be computed for each new candidate cluster, the overall complexity of Step 5 would be $\mathcal{O}(|\mathcal{X}||\mathcal{N}|k + |\mathcal{S}|mk)$. Step 6 is in $\mathcal{O}(|\mathcal{X}||\mathcal{N}|)$ whereas Step 7 is simply in $\mathcal{O}(|\mathcal{S}|m)$. With the above steps having to be performed for each $X$ and for each iteration, the overall {\it FairKM} complexity works out to be in $\mathcal{O}(|\mathcal{X}|^2|\mathcal{N}|kl + |\mathcal{X}||\mathcal{S}|mkl)$ where $l$ is the number of iterations. While the quadratic dependency on the dataset size makes {\it FairKM} much slower than simple $K$-Means (which is linear on dataset size), {\it FairKM} compares very favorably against other fair clustering methods (e.g., exact fairlet decomposition~\cite{chierichetti2017fair} is NP-hard, and even the proposed approximation is super-quadratic) which are computationally intensive. 

\subsection{FairKM Extensions}

We outline two extensions to the basic {\it FairKM} outlined earlier which was intended towards handling numeric non-sensitive attributes and multi-valued sensitive attributes. 

\subsubsection{Extension to Numeric Sensitive Attributes}

{\it FairKM} is easily adaptable to numeric sensitive attributes (e.g., {\it age} for cases where that is appropriate). If all attributes in $\mathcal{S}$ are numeric, the fairness loss term in Eq.~\ref{eq:fairnessterm} would be written out as:

\begin{multline}
    deviation_{\mathcal{S}}(\mathcal{C},\mathcal{X}) = 
    \sum_{C \in \mathcal{C}} \bigg( \frac{|C|}{|\mathcal{X}|} \bigg)^2 \times \sum_{S \in \mathcal{S}} (C.S - \mathcal{X}.S)^2
    \label{eq:fairnessterm1}
\end{multline}

where $C.S$ and $\mathcal{X}.S$ indicate the average value of the numeric attribute $S$ across objects in $C$ and $\mathcal{X}$ respectively. When there are a mix of multi-valued and numeric attributes, the inner term would take the form of Eq.~\ref{eq:fairnessterm} and Eq.~\ref{eq:fairnessterm1} for multi-valued and numeric attributes respectively. These entail corresponding changes to the update equations which we do not describe here for brevity. 

\subsubsection{Extension to allow Sensitive Attribute Weighting}

In certain scenarios, some sensitive attributes may need to be considered more important than others. This may be due to historical reasons based on a legacy of documented high discrimination on certain attributes, or due to visibility reasons where discrimination on certain attributes (e.g., {\it gender}, {\it race} and {\it sexual orientation}) being more visible than others (e.g., {\it country of origin}). The {\it FairKM} framework could easily be extended to allow for differential attribute-specific weighting by changing the deviation term to be as follows:

\begin{multline}
    deviation_{\mathcal{S}}(\mathcal{C},\mathcal{X}) = \\
    \sum_{C \in \mathcal{C}} \bigg( \frac{|C|}{|\mathcal{X}|} \bigg)^2 \times \sum_{S \in \mathcal{S}} w_S \times \frac{\sum_{s \in Values(S)} \big( Fr_C^S(s) - Fr_{\mathcal{X}}^S(s) \big)^2}{|Values(S)|}
    \label{eq:fairnessterm2}
\end{multline}

Attributes that are more important for fairness considerations can then be assigned a higher weight, i.e. $w_S$, which would lead to their loss being amplified, thus incentivizing {\it FairKM} to focus more on them for fairness, consequently leading to a higher representational fairness over them, within the clusters in the output. The $w_S$ terms would then also affect the update equations. 

\begin{table*}
\begin{tabular}{ll}
\\
\cline{1-2}
{\bf Type} & {\bf Description} \\
\cline{1-2}
{\bf 1:}Horizontal Motion & The object involved is in a horizontal straight line motion. \\
{\bf 2:}Vertical motion with an initial velocity & The object is thrown straight up or down with a velocity. \\
{\bf 3:}Free fall & The object is in a free fall. \\
{\bf 4:}Horizontally projected & The object is projected horizontally from a height. \\
{\bf 5:}Two-dimensional & The body is projected with a velocity at an angle to the horizontal.\\
\hline
\cline{1-2}
\end{tabular}
    \caption{Kinematics Word Problem Types} 
\label{tab:wptypes}
\end{table*}

\section{Experimental Study}

We now detail our experimental study to gauge and quantify the effectiveness of {\it FairKM} in delivering good quality and fair clusterings against state-of-the-art baselines. We first outline the datasets in our experimental setup, followed by a description of the evaluation measures and baselines. This is then followed by our results and an analysis of the results. 

\begin{table}
\begin{tabular}{ll}
\\
\cline{1-2}
{\bf Attribute} & {\bf No. of values} \\
\cline{1-2}
Marital status & 7  \\
Relationship status& 6 \\
Race & 5 \\
Gender & 2 \\
Native country& 41\\
\cline{1-2}
\end{tabular}
    \caption{Adult Dataset: Number of possible values for each sensitive attribute} 
\label{tab:stat1}
\end{table}

\subsection{Datasets}


We use two real-world datasets in our empirical study. The datasets are chosen to cover very different domains, attributes and dataset sizes, to draw generalizable insights from the study. First, we use the popular \textit{Adult} dataset from UCI repository~\cite{dua2017uci}; this dataset is sometimes referenced as the {\it Census Income} dataset and contains information from the 1994 US Census. The dataset  has $32561$ instances, each instance represented using $13$ attributes. Among the $13$ attributes, $5$ are chosen to form the set of sensitive attributes, $\mathcal{S}$. These are \{$marital\ status$, $relationship$ $status$, $race$, $gender$, $native$ $country$\}. The number of values taken by each of the sensitive attributes are shown in Table \ref{tab:stat1}. The set of non-sensitive attributes, $\mathcal{N}$, pertain to {\it age}, {\it work class} (2 attributes), {\it education} (2 attributes), {\it occupation}, {\it fiscal information} (2 attributes) and {\it number of working hours}. The dataset has been widely used for predicting income as belonging to one of $>50k\$$ or $<=50k\$$. We first undersample the dataset to ensure parity across this income class attribute that we do not use in the clustering process. The total number of instances after undersampling is $15682$. Second, we use a dataset\footnote{https://github.com/savithaabraham/Datasets} of $161$ word problems from the domain of kinematics. Kinematics is the study of motion without considering the cause of motion. The problems in this dataset is categorized into various types as indicated in Table~\ref{tab:wptypes}. The complexity of a word problem typically depends on the type. For example, Type 1 problems are easier to solve (in terms of the equations required) compared to Type 5 problems. Table \ref{tab:stat2} shows the number of problems of each of the above types in the dataset. Given such a dataset of word problems from kinematics domain, we are interested in the task of clustering the word problems such that the proportional representation of problems of a particular type in a cluster reflects its representation in the entire dataset. In the application scenario of automatic construction of multiple questionnaires (one from each cluster) from a question bank, the fair clustering task corresponds to ensuring that each questionnaire contains a reasonable mix of problem types. This ensures that there is minimal asymmetry between the different questionnaires generated by a clustering, in terms of overall hardness. For the fair clustering formulation, thus, the problem types form the set of $5$ sensitive binary attributes, $\mathcal{S}$. The lexical representation of each word problem, as a $100$ dimensional vector using Doc2Vec models~\cite{doc2vec}, forms the set of numeric attributes in $\mathcal{N}$. Given our fairness consideration, we consider achieving a fair proportion of word problem types within each cluster that reflects their proportion across the dataset. 

It may be noted that the {\it Adult} and {\it Kinematics} datasets come from different domains (Census and Word Problems/NLP respectively), have different sizes of non-sensitive attribute sets (8 and 100 attributes in $\mathcal{N}$ respectively), different kinds of sensitive attribute sets (multi-valued and binary respectively) and have widely varying sizes ($15k$ and $161$ respectively). An empirical evaluation over such widely varying datasets, we expect, would inspire confidence in the generalizability of empirical results. 


\begin{table}[]
\begin{tabular}{ll}
\cline{1-2}
{\bf Type} & {\bf Count} \\
\cline{1-2}
1 - Horizontal motion & 60  \\
2 - Vertical motion with an initial velocity & 36 \\
3 - Free fall & 15 \\
4 - Horizontally projected  & 31 \\
5 - Two-dimensional & 19\\
\cline{1-2}
\end{tabular}
    \caption{Kinematics Dataset: \#Problems of each Type} 
\label{tab:stat2}
\end{table}
  
\subsection{Evaluation}\label{sec:evalmeasures}

Having defined the task of fair clustering in Section~\ref{sec:probdef}, it follows that a fair clustering algorithm would be expected to perform well on two sets of evaluation metrics, those that relate to clustering quality over $\mathcal{N}$ and those that relate to fairness over $\mathcal{S}$. We now outline such evaluation measures below, in separate subsections. 

\subsubsection{Clustering Quality}\label{sec:clusteringquality}

These measure how well the clustering fares in generating clusters that are coherent on attributes in $\mathcal{N}$, and do not depend on attributes in $\mathcal{S}$. These could include:

\begin{itemize}[leftmargin=*]
\item {\it Silhouette Score} {\bf (SH):} Silhouette~\cite{rousseeuw1987silhouettes} measures the separatedness of clusters, and quantifies a clustering with a score in $[-1,+1]$, higher values indicating well-separated clusters. 
\item {\it Clustering Objective} {\bf (CO):} Clustering objective functions such as those employed by $K$-Means~\cite{macqueen1967some} measure how much observations deviate from the centroids of the clusters they are assigned to, where lower values indicate coherent clusters. In particular, the $K$-Means objective function is:
\begin{equation}
\sum_{C \in \mathcal{C}} \sum_{X \in C} dist_{\mathcal{N}}(X,C)
\end{equation}
where $C$ stands for both a cluster in the clustering $\mathcal{C}$ as well as the prototype object for the cluster, and $dist_{\mathcal{N}}(.,.)$ is the distance measure computed over attributes in $\mathcal{N}$.
\item {\it Deviation from $\mathcal{S}$-blind Clusterings:} $\mathcal{S}$-blind clusterings may be thought of achieving the best possible clusters for the task when no fairness considerations are imposed. Thus, among two clusterings of similar fairness, that with lower deviation from $\mathcal{S}$-blind clusterings may be considered desirable. A fair clustering can be compared with a $\mathcal{S}$-blind clustering using the following two measures:
\begin{itemize}[leftmargin=*]
\item {\it Centroid-based Deviation} {\bf (DevC):} Consider each clustering to be represented as a set of cluster centroids, one for each cluster within the clustering. The sum of pair-wise dot-products between centroid pairs, each pair constructed using one centroid from the fair clustering and one from the $\mathcal{S}$-blind clustering, would be a measure of deviation between the clusterings. Such measures have been used in generating disparate clusterings~\cite{jain2008simultaneous}. 
\item {\it Object pair-wise Deviation} {\bf (DevO):} Consider each pair of objects from $\mathcal{X}$, and one clustering (either of $\mathcal{S}$-blind and fair); the objects may belong to either the {\it same} cluster or to {\it different} clusters. The fraction of object pairs from $\mathcal{X}$ where the same/different verdicts from the two clusterings disagree provide an intuitive measure of deviation between clusterings. 
\end{itemize}
\end{itemize}

\subsubsection{Fairness} 

These measure the fairness of the clustering output from the (fair) clustering algorithm. Analogous to clustering quality measures that depend only on $\mathcal{N}$, the fairness measures we outline below depend only on $\mathcal{S}$ and are independent of $\mathcal{N}$. As outlined earlier, we quantify unfairness as the extent of deviation between representations of groups defined using attributes in $\mathcal{S}$ in the dataset and each cluster in the clustering output. Consider a multi-valued attribute $S \in \mathcal{S}$, which can take on $t$ values. The normalized distribution of presence of each of the $t$ values in $\mathcal{X}$ yields a t-length probability distribution vector $\mathcal{X}_S$. A similar probability distribution can then be computed for each cluster $C$ in the clustering $\mathcal{C}$, denoted $C_S$. Different ways of measuring the cluster-specific deviations $\{ \ldots, dev(C_S,\mathcal{X}_S), \ldots \}$ and aggregating them to a single number yield different quantifications of fairness, as below:
\begin{itemize}[leftmargin=*]
\item {\it Average Euclidean} {\bf (AE):} This measures the average of cluster-level deviations, deviations quantified using euclidean distance between representation vectors (i.e., $\mathcal{X}_S$ and $C_S$s). Since clusters may not always be of uniform sizes, we use a cluster-cardinality weighted average. 
\begin{equation}
AE_S = \frac{\sum_{C \in \mathcal{C}} |C| \times ED(C_S,\mathcal{X}_S)}{\sum_{C \in \mathcal{C}}|C|}
\end{equation}
where $ED(.,.)$ denotes the euclidean distance.
\item {\it Average Wasserstein} {\bf (AW):} In this measure, the deviation is computed using Wasserstein distance in lieu of Euclidean, as used in~\cite{wang2019towards}, with other aspects remaining the same as above. 
\item {\it Max Euclidean} {\bf (ME):} Often, just optimizing for average fairness across clusters is not enough since there could be a very skewed (small) cluster, whose effect may be obscured by other clusters. It is often the case that one or few clusters get picked from a clustering to be actioned upon. Thus, the maximum skew is of interest as an indicative upper bound on the unfairness the clustering could cause if {\it any} one of its clusters is chosen for further action. 
\item {\it Max Wasserstein} {\bf (MW):} This uses Wasserstein instead of Euclidean, using the same formulation as {\it Max Euclidean}. 
\end{itemize}
When there are multiple attributes in $\mathcal{S}$, as is often the case, the average of the above measures across attributes in $\mathcal{S}$ provides aggregate quantifications. As may be evident, the above constructions work only for categorical attributes; however, a similar set of measures can be readily devised for numeric attributes in $\mathcal{S}$. With our datasets containing only categorical attributes among $\mathcal{S}$, we do not outline the corresponding metrics for numeric attributes, though they follow naturally. We are unable to apply some popular fairness evaluation metrics such as {\it balance}~\cite{chierichetti2017fair} due to them being devised for binary attributes.  

\subsection{Baselines}
We compare our approach against two baselines. The first is that of $\mathcal{S}$-blind $K$-Means clustering, that performs $K$-Means clustering on data using the attributes in $\mathcal{N}$ alone. This baseline is code-named $K$-Means ($\mathcal{N}$). $K$-Means ($\mathcal{N}$) will produce the most coherent clusters on $\mathcal{N}$ as its objective function just focuses on maximizing intra-cluster similarity and minimizing inter-cluster similarity over $\mathcal{N}$, unlike $FairKM$ that has an additional fairness constraint which may result in compromising the coherence goal. Comparing the two enables us to evaluate the extent to which cluster coherence is traded off by $FairKM$ in generating fairer clusters. The second baseline is the approach described in \cite{ziko2019clustering} which is a fair version of $K$-Means clustering for scenarios involving a single multi-valued sensitive attribute. We will refer to this baseline as {\it ZGYA} from here, based on the names of the authors. Since it is designed for a single multi-valued sensitive attribute and cannot handle multiple sensitive attributes within its formulation, we invoke {\it ZGYA} multiple times, separately for each attribute in $\mathcal{S}$. Each invocation is code-named $ZGYA(S)$ where $S$ is the sensitive attribute used in the invocation. We also report results for similar runs of $FairKM$, where we consider just one of the attributes in $\mathcal{S}$ as sensitive at a time. The comparative evaluation between {\it FairKM} and {\it ZGYA} enables studying the effectiveness of FairKM formulation over that of {\it ZGYA} in their relative effectiveness of trading off coherence for fairness. 

\subsection{Setting $\lambda$ in {\it FairKM}} \label{setlambda}

From Eq.~\ref{eq:objective}, it may be seen that the $K$-Means term has a contribution from each object in $\mathcal{X}$, whereas the fairness term (Eq.~\ref{eq:fairnessterm}) aggregates cluster level contributions. This brings a disparity in that the former has $|\mathcal{X}|/k$ times as many terms as the latter. Further, it may be noted that the fairness term aggregates deviations between cluster level fractional representations and dataset level fractional representation. The fractional representation being an average across objects in the cluster, each object can only influence $1/|C|$ of it, where $|C|$ is the cluster cardinality. On an average, across clusters, $|C| = |\mathcal{X}|/k$. Thus, the fairness term has $|\mathcal{X}|/k$ fewer terms, each of whom can be influenced by an object to a fraction of $1/(|\mathcal{X}|/k)$. To ensure that the terms are of reasonably similar sizes, so the clustering quality and fairness concerns be given similar weighting, the above observations suggest that $\lambda$ be set to $\big(\frac{|\mathcal{X}|}{k}\big)^2$. From our empirical observations, we have seen that the {\it FairKM} behavior varies smoothly around this setting. Based on the above heuristic, we set $\lambda$ to $10^6$ for the Adult dataset, and $10^3$ for the Kinematics dataset, given their respective sizes. We will empirically analyze sensitivity to $\lambda$ in Section~\ref{sec:lambdasensitivity}. We set max iterations to $30$ in {\it FairKM} instantiations.


\begin{table*}[]
\small
\centering
\resizebox{0.8\textwidth}{!}{
\begin{tabular}{|p{1.2 cm}|p{1.6 cm}p{1.5 cm}p{1.5 cm}||p{1.6 cm}p{1.5 cm}p{1.5 cm}|}
\cline{1-7}
Evaluation & \multicolumn{3}{c||}{k=$5$}&  \multicolumn{3}{c|}{k=$15$} \\
\cline{2-4}
\cline{5-7}
Measure& \multicolumn{1}{|l|}{$K$-Means ($\mathcal{N}$)} & Avg. ZGYA & FairKM & \multicolumn{1}{|l|}{$K$-Means ($\mathcal{N}$)} & Avg. ZGYA & FairKM   \\
\cline{1-7}
CO $\downarrow$ &	\multicolumn{1}{|l|}{1120.9112}&	10791.8311	& {\bf 1345.1688}	&\multicolumn{1}{|l|}{837.9785}	&4095.8366&	{\bf 1235.2859} \\
SH $\uparrow$ &	\multicolumn{1}{|l|}{0.7212}	&0.0557&	{\bf 0.3918}	&\multicolumn{1}{|l|}{0.6076}&	0.0573&	{\bf 0.3747}\\
DevC $\downarrow$ &	\multicolumn{1}{|l|}{0.0}&	{\bf 8.4597}	&8.4707&	\multicolumn{1}{|l|}{0.0}&	39.3615&	{\bf 13.1244}\\
DevO $\downarrow$ &	\multicolumn{1}{|l|}{0.0}&	0.0306&	{\bf 0.0233}&	\multicolumn{1}{|l|}{0.0}&	0.0360	&{\bf 0.0256}\\
\cline{1-7}
\end{tabular}
}
\caption{Clustering quality on Adult Dataset -  FairKM vs. Average across $\{ZGYA(S)|S \in \mathcal{S}\}$, shown with $K$-Means$(\mathcal{N})$.} 
\label{tab:res1}
\end{table*}

\begin{table*}
\resizebox{\textwidth}{!}{%
\begin{tabular}{|p{2.0 cm}p{2.0 cm}||p{1.8 cm}p{1.8 cm}p{1.8 cm}p{1.8 cm}||p{1.8 cm}p{1.8 cm}p{1.8 cm}p{1.8 cm}|}
\cline{1-10}
\cline{1-10}
$\mathcal{S}$& Evaluation & \multicolumn{4}{c||}{k=$5$} & \multicolumn{4}{c|}{k=$15$} \\
\cline{3-6}
\cline{7-10}
Attribute & Measure&$K$-Means($\mathcal{N}$)& ZGYA(S)&FairKM&FairKM Impr(\%)&$K$-Means($\mathcal{N}$)&ZGYA(S)&FairKM&FairKM Impr(\%)\\
\cline{1-10} 
\cline{1-10}
{\bf Mean}&AE&	0.0459&	0.1201&	\textbf{0.0278}&	\textbf{39.5357}&	0.0537&	0.1289&	\textbf{0.0295}&	\textbf{45.0796}\\
{\bf across $\mathcal{S}$}&AW	&0.0161&	0.0370&	\textbf{0.0087}&	\textbf{45.7857}&	0.0194&	0.0398&	\textbf{0.0094}&	\textbf{51.7043}\\
{\bf Attributes} &ME	&0.2063&	0.8729&	\textbf{0.1457}&	\textbf{29.4002}&	0.2475&	0.7810&	\textbf{0.1542}&	\textbf{37.6985}\\
&MW	&0.0740&	0.1235&	\textbf{0.0502}&	\textbf{32.0985}&	0.0753&	0.1262&	\textbf{0.0542}&	\textbf{28.0040}\\
\cline{1-10}
\cline{1-10}
\multicolumn{10}{|c|}{Results for Each Sensitive Attribute in $\mathcal{S}$ below.} \\
\cline{1-10}
Marital Status &AE	&0.0792&	0.0886&	\textbf{0.0539}&	\textbf{31.9408}&	0.0853&	0.1318&	\textbf{0.0558}&	\textbf{34.5263}\\
&AW&	0.0182&	0.0159&	\textbf{0.0132}	&\textbf{16.5650}&	0.0191&	0.0258	&\textbf{0.0136}&	\textbf{28.4239}\\
&ME	&0.3055&	0.7356	&\textbf{0.2578}&	\textbf{15.6087}&	0.3572	&0.6365&	\textbf{0.2607}	&\textbf{27.0042}\\
&MW	&\textbf{0.0573}&	0.0890	&0.0592&	-3.3881&	\textbf{0.0566}	&0.0952&	0.0604	&-6.6317\\
\cline{1-10}
Rel. Status&AE	&0.0711&	0.1743&	\textbf{0.0486}&	\textbf{31.5656}	&0.0808&	0.1903&	\textbf{0.0500}&	\textbf{38.1517}\\
&AW	&0.0197	&0.0371&	\textbf{0.0146}&	\textbf{25.8744}&	0.0219&	0.0429&	\textbf{0.0150}	&\textbf{31.3346}\\
&ME	&0.3331	&0.7796&	\textbf{0.2717}&	\textbf{18.4487}&	0.3823&	0.7804&	\textbf{0.2777}	&\textbf{27.3667}\\
&MW	&\textbf{0.0732}	&0.1205&	0.0760&	-3.8026&	\textbf{0.0750}&	0.1439&	0.0776	&-3.4770\\
\cline{1-10}
Race&AE&	0.0163&	0.0564&	\textbf{0.0066}&	\textbf{59.2251}&	0.0168&	0.0647&	\textbf{0.0079}&	\textbf{53.0164}\\
&AW	&0.0053&	0.0154&	\textbf{0.0023}&	\textbf{55.9473}&	0.0055&	0.0162&	\textbf{0.0028}&	\textbf{48.9813}\\
&ME	&0.0385&	1.0085&	\textbf{0.0266}&	\textbf{30.8822}&	0.0565&	1.2175&	\textbf{0.0336}&	\textbf{40.6276}\\
&MW	&0.0126&	0.1159&	\textbf{0.0092}&	\textbf{27.3039}&	0.0165&	0.1142& \textbf{0.0115}& \textbf{30.2523}\\
\cline{1-10}
Gender&AE&	0.0529&	0.2535&	\textbf{0.0183}&	\textbf{65.3039}&	0.0711&	0.2256&	\textbf{0.0208}&	\textbf{70.7472}\\
&AW	&0.0370&	0.1153&	\textbf{0.0130}&	\textbf{64.9210}&	0.0499&	0.1122&	\textbf{0.0147}&	\textbf{70.4913}\\
&ME	&0.3324&	0.9793&	\textbf{0.1487}&	\textbf{55.2713}&	0.4028&	1.0201&	\textbf{0.1697}&	\textbf{57.8731}\\
&MW	&0.2254&	0.2568&	\textbf{0.1051}&	\textbf{53.3681}&	0.2262&	0.2671&	\textbf{0.1200}&	\textbf{46.9680}\\
\cline{1-10}
Native Country&AE&	\textbf{0.0101}&	0.0276&	0.0113&	-11.2331&	0.0146&	0.0323&	\textbf{0.0130}&	\textbf{10.9108}\\
&AW	&\textbf{0.0005}&	0.0013&	0.0006&	-15.2027&	0.0007&	0.0015&	\textbf{0.0006}&	\textbf{4.7201}\\
&ME	&\textbf{0.0221}&	0.8612&	0.0236&	-6.4585&	0.0385&	0.2506&	\textbf{0.0292}&	\textbf{24.1555}\\
&MW	&\textbf{0.0012}&	0.0354&	0.0016&	-25.8608&	0.0020&	0.0107&	\textbf{0.0015}&	\textbf{25.6292}\\
\cline{1-10}

\end{tabular}%
}
\caption{Fairness evaluation on Adult Dataset - $\mathcal{S}$-blind $K$-Means, Single invocation of {\it FairKM} on all $\mathcal{S}$ attributes, Separate Invocations of ZGYA on each attribute in $\mathcal{S}$. (Note: This is a synthetic favorable setting for ZGYA, to stress test {\it FairKM} against ZGYA).}
\label{tab:res3}
\end{table*}

\begin{table}[]
\small
\centering
\resizebox{0.5\textwidth}{!}{
\begin{tabular}{|p{1.2 cm}|p{1.6 cm}p{1.5 cm}p{1.5 cm}|}
\cline{1-4}
Evaluation& \multicolumn{1}{|l|}{$K$-Means ($\mathcal{N}$)} & Avg. ZGYA & FairKM \\
\cline{1-4}
CO $\downarrow$	&\multicolumn{1}{|l|}{145.6441}	&164.4703&	{\bf 148.1003}\\
SH $\uparrow$	&\multicolumn{1}{|l|}{0.0390}&	-0.0001	&{\bf 0.0149}\\
DevC $\downarrow$	&\multicolumn{1}{|l|}{0.0}&	1.1844	&{\bf 1.1241}\\
DevO $\downarrow$	&\multicolumn{1}{|l|}{0.0}&{\bf 0.0032}&	0.0038\\
\cline{1-4}
\end{tabular}
}
\caption{Clustering quality on Kinematics Dataset -  {\it FairKM} vs. Average across $\{ZGYA(S)|S \in \mathcal{S}\}$, shown with $K$-Means$(\mathcal{N})$.} 
\label{tab:res2}
\end{table}

\setlength{\tabcolsep}{5pt}
\begin{table}
\resizebox{0.5\textwidth}{!}{%
\begin{tabular}{|p{1.5 cm}p{1.0 cm}|p{1.4 cm}p{1.4 cm}p{1.4 cm}p{1.4 cm}|}
\cline{1-6}
$\mathcal{S}$ Attribute &Metric&$K$-Means ($\mathcal{N}$)&ZGYA (S)&FairKM&FairKM Impr(\%)\\
\cline{1-6}
{\bf Mean}&AE	&0.1704	&0.1183	&\textbf{0.0172}	&\textbf{85.4311}\\
{\bf across $\mathcal{S}$}&AW	&0.1021	&0.0766	&\textbf{0.0120}	&\textbf{84.3660}\\
{\bf Attributes}&ME	&0.3744	&0.2571	&\textbf{0.1488}	&\textbf{42.1364}\\
&MW	&0.2083	&0.1676	&\textbf{0.0852}	&\textbf{49.1420}\\
\cline{1-6}
\multicolumn{6}{|c|}{Results for Each Sensitive Attribute in $\mathcal{S}$ below.} \\
\cline{1-6}
Type-1 &AE	&0.2567	&0.1821	&\textbf{0.0148}	&\textbf{91.8775}\\
&AW	&0.1289	&0.1000	&\textbf{0.0103}	&\textbf{89.7246}\\
&ME	&0.4909	&0.3502	&\textbf{0.1673}	&\textbf{52.2397}\\
&MW	&0.2828	&0.2321	&\textbf{0.1004}	&\textbf{56.7159}\\
\cline{1-6}
Type-2&AE	&0.2145	&0.1481	&\textbf{0.0163}	&\textbf{88.9722}\\
&AW	&0.1213	&0.0994	&\textbf{0.0113}	&\textbf{88.6729}\\
&ME	&0.5116	&0.3398	&\textbf{0.1600}	&\textbf{52.9166}\\
&MW	&0.2149	&0.1931	&\textbf{0.0888}	&\textbf{54.0235}\\
\cline{1-6}
Type-3&AE	&0.0759	&0.0604	&\textbf{0.0178}	&\textbf{70.5473}\\
&AW	&0.0535	&0.0427	&\textbf{0.0123}	&\textbf{71.2578}\\
&ME	&0.1935	&\textbf{0.1270}	&0.1527	&-20.2176\\
&MW	&0.1206	&0.0898	&\textbf{0.0754}	&\textbf{16.0235}\\
\cline{1-6}
Type-4&AE	&0.1631	&0.1009	&\textbf{0.0152}	&\textbf{84.9649}\\
&AW	&0.1079	&0.0708	&\textbf{0.0107}	&\textbf{84.9541}\\
&ME	&0.3605	&0.2410	&\textbf{0.1263}	&\textbf{47.5836}\\
&MW	&0.2103	&0.1662	&\textbf{0.0770}	&\textbf{53.6570}\\
\cline{1-6}
Type-5&AE	&0.1415	&0.0999	&\textbf{0.0221}	&\textbf{77.8973}\\
&AW	&0.0989	&0.0703	&\textbf{0.0154}	&\textbf{78.0243}\\
&ME	&0.3155	&0.2273	&\textbf{0.1375}	&\textbf{39.5175}\\
&MW	&0.2128	&0.1569	&\textbf{0.0846}	&\textbf{46.1075}\\
\cline{1-6}
\end{tabular}%
}
\caption{Fairness evaluation on Kinematics Dataset - $\mathcal{S}$-blind $K$-Means, Single invocation of {\it FairKM} on all $\mathcal{S}$ attributes, Separate Invocations of ZGYA on each attribute in $\mathcal{S}$. (Note: This is a synthetic favorable setting for ZGYA, to stress test {\it FairKM} against ZGYA).}
\label{tab:res4}
\end{table}

\subsection{Clustering Quality and Fairness}

\subsubsection{Evaluation Setup}

In each of our datasets, there are five sensitive (i.e., $\mathcal{S}$) attributes. {\it FairKM} can be instantiated with all of them at once, and we do so with appropriate values of $\lambda$ ($10^6$ or $10^3$, as mentioned in Section \ref{setlambda}). We perform $100$ such instantiations, each with a different random seed, and measure the {\it clustering quality} and {\it fairness} evaluation measures (fairness measures computed separately for each attribute in $\mathcal{S}$ as well as the average across all attributes in $\mathcal{S}$) outlined in Section~\ref{sec:evalmeasures}. We take the mean values across the $100$ instantiations to arrive at a single robust value for each evaluation measure for {\it FairKM}. An analogous setting is used for our first baseline, the $\mathcal{S}$-blind $K$-Means (denoted $K$-Means $(\mathcal{N})$), as well. Our second baseline, {\it ZGYA}, unlike {\it FairKM}, needs to be instantiated with one $\mathcal{S}$ attribute at a time. Given this setting, we adopt different mechanisms to compare {\it FairKM} against $ZGYA$ across clustering quality and fairness evaluation measures. First, for {\it clustering quality}, we instantiate $ZGYA$ separately with each attribute in $\mathcal{S}$ and compute an average value for each evaluation measure across random initializations as described previously. This yields one value for each evaluation measure for each attribute in $\mathcal{S}$, which we take an average of, and report as the {\it clustering quality} of {\it Avg. ZGYA}. Second, for {\it fairness}, we adopt a synthetic favorable setting for {\it ZGYA} to test {\it FairKM} against. For each attribute $S \in \mathcal{S}$, we consider the fairness metrics (AE, AW, ME, MW) obtained by the instantiation of {\it ZGYA} over {\sl only} that attribute (averaged across random initializations, as earlier). This is compared to the fairness metrics obtained for $S$ by the {\it FairKM} instantiation that considers {\sl all} attributes in $\mathcal{S}$. In other words, for each $S \in \mathcal{S}$, we benchmark the single cross-$\mathcal{S}$ instantiation of {\it FairKM} against separate $S$-targeted instantiations of {\it ZGYA}. We also report the average across these separate comparisons across attributes in $\mathcal{S}$. For $K$-Means style clustering formulations, the number of clusters $k$, is an important parameter. We experiment with two values for $k$, viz., $5$ and $15$, for the Adult dataset, whereas we use $k=5$ for the Kinematics dataset, given its much smaller size. 

\subsubsection{Clustering Quality}

The clustering quality results appear in Table~\ref{tab:res1} (Adult dataset) and Table~\ref{tab:res2} (Kinematics dataset), with the direction against each evaluation measure indicating whether lower or higher values are more desirable. For clustering quality metrics that depend only on attributes in $\mathcal{N}$, we use $K$-Means ($\mathcal{N}$) as a reference point since that is expected to perform well, given that it does not need to heed to $\mathcal{S}$ and is not held accountable for fairness. Thus, {\it FairKM} is not expected to beat $K$-Means ($\mathcal{N}$); the lesser the degradation from $K$-Means ($\mathcal{N}$) on various clustering quality metrics, the better it may be regarded to be. We compare {\it FairKM} and {\it Avg. ZGYA} across the results tables, highlighting the better performer on each evaluation measure by boldfacing the appropriate value. On the Adult dataset (Table~\ref{tab:res1}), it may be seen that {\it FairKM} performs better than {\it Avg. ZGYA} on seven out of eight combinations, with it being competitive with the latter on the eighth. {\it FairKM} is seen to score significantly better than {\it Avg. ZGYA} on clustering objective (CO) and silhoutte score (SH), with the gains on the deviation metrics (DevC and DevO) being more modest. It may be noted that $CO$ and $SH$ may be regarded as more reliable measures, since they evaluate the clustering directly. In contrast, $DevO$ and $DevC$ evaluate the deviation against reference $K$-Means $(\mathcal{N})$ clusterings; these deviation measures penalize deviations even if those be towards {\it other} good quality clusterings that may exist in the dataset. The trends from the Adult dataset hold good for the Kinematics dataset as well (see Table~\ref{tab:res2}), confirming that the trends generalize well across datasets of widely varying character. Overall, our results indicate that {\it FairKM} is able to generate much better quality clusterings than {\it Avg. ZGYA}, when gauged on attributes in $\mathcal{N}$. 

\subsubsection{Fairness}

The fairness evaluation measures for the Adult and Kinematics datasets appear in Tables~\ref{tab:res3} and~\ref{tab:res4} respectively; it may be noted that lower values are desirable on all evaluation measures, given that they all measure deviations. In these results, which include a synthetically favorable setting for {\it ZGYA} (as noted earlier), the top block indicates the average results across all attributes in $\mathcal{S}$, with the following result blocks detailing the results for the specific parameters in $\mathcal{S}$. The overarching summary of this evaluation suggests, as indicated in the top-blocks across the two tables, that {\it FairKM} surpasses the baselines with significant margins. The $\%\ impr$ column indicates the gain achieved by {\it FairKM} over the next best competitor. The percentage improvements recorded are around $35+\%$ on an average for the Adult dataset, whereas the corresponding figure is higher, at around $60+\%$ for the Kinematics dataset. We wish to specifically make a few observations from the results. {\it First}, the closest competitor to {\it FairKM} is {\it ZGYA} on the Kinematics dataset, whereas $K$-Means ($\mathcal{N}$) curiously outperforms {\it ZGYA} quite consistently on the Adult dataset. This indicates that {\it ZGYA} is likely more suited to settings where the number of values taken by the sensitive attribute is less. In our case, the Kinematics dataset has all binary attributes in $\mathcal{S}$, whereas Adult dataset has sensitive attributes that take as many as $41$ different values. {\it Second}, while {\it FairKM} is designed to accommodate $\mathcal{S}$ attributes that take many different values, the fairness deviations appear to degrade, albeit at a much lower pace than {\it ZGYA}, as attributes take on very many values. This is indicated by the lower performance (with small margins) on the {\it native country} (41 values) attribute at $k=5$ in Table~\ref{tab:res3}. However, promisingly, it is able to utilize the additional flexibility that is provided by larger $k$s to ensure higher rates of fairness on them. As may be seen, {\it FairKM} recovers well to perform significantly better on {\it native country} at $k=15$. This indicates that {\it FairKM} will benefit from a higher flexibility in cluster assignment (with higher $k$) when there are a number of (high cardinality) attributes to ensure fairness over. {\it Third}, the {\it FairKM} formulation targets to minimize overall fairness and does not specifically nudge it towards ensuring good performance on the {\it max} measures (ME and MW) that quantify the worst deviation across clusters. Thus, it's design allows to choose higher fairness in multiple clusters even at the expense of disadvantaging fairness in one or few clusters, which is indeed undesirable. The performance on ME and MW suggest that such trends are not widely prevalent, with {\it FairKM} recording reasonable gains on ME and ME. However, cases such as {\it marital status} in Table~\ref{tab:res3} and Type-3 in Table~\ref{tab:res4} suggest that is a direction in which {\it FairKM} could improve. {\it Finally}, the overall summary from Tables~\ref{tab:res3} and~\ref{tab:res4} suggest that {\it FairKM} delivers much fairer clusters on $\mathcal{S}$ attributes, and records significant gains over the baselines, in our empirical evaluation. 

\begin{figure}[tb]
\centering
\includegraphics[width=\columnwidth]{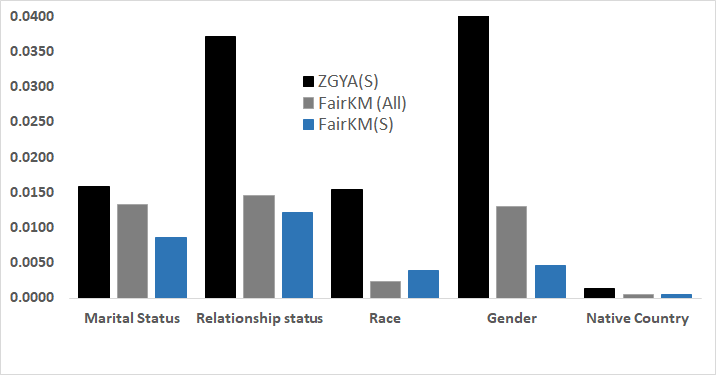}
\caption{Adult Dataset: AW Comparison}
\label{fig:adultaw}
\end{figure}

\begin{figure}[tb]
\centering
\includegraphics[width=\columnwidth]{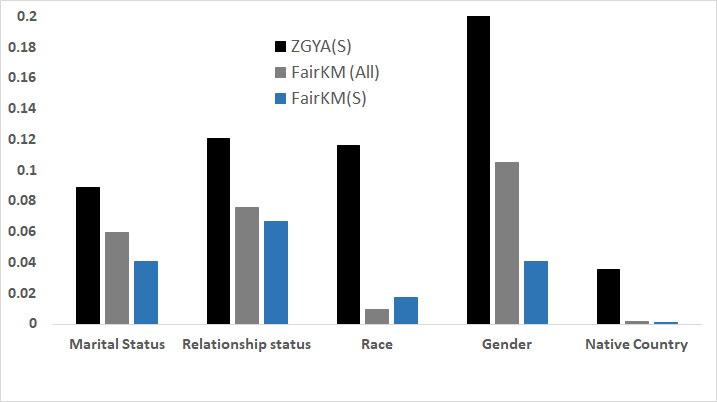}
\caption{Adult Dataset: MW Comparison}
\label{fig:adultmw}
\end{figure}

\begin{figure}[tb]
\centering
\includegraphics[width=\columnwidth]{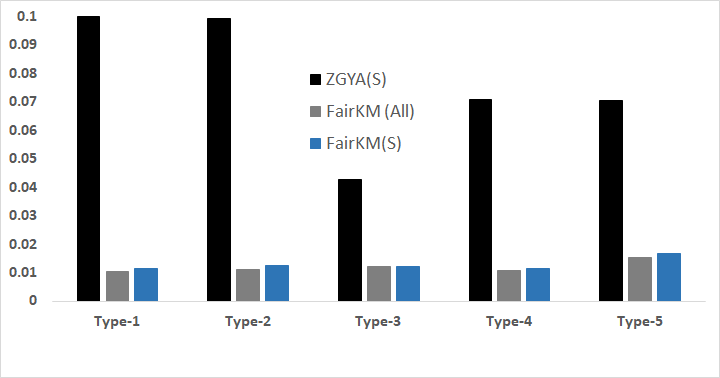}
\caption{Kinematics Dataset: AW Comparison}
\label{fig:kinematicsaw}
\end{figure}

\begin{figure}[tb]
\centering
\includegraphics[width=\columnwidth]{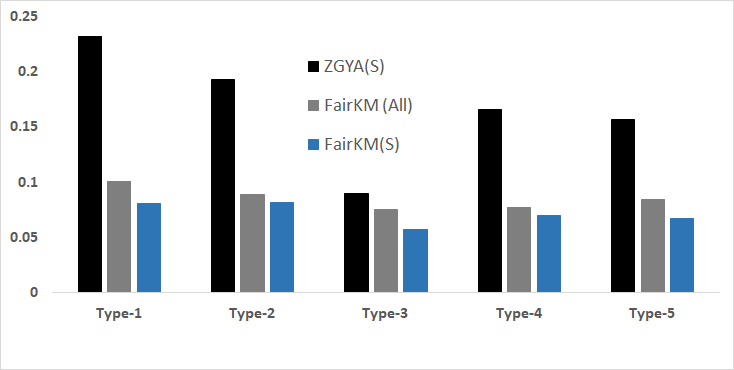}
\caption{Kinematics Dataset: MW Comparison}
\label{fig:kinematicsmw}
\end{figure}

\subsection{{\it FairKM} vs. {\it ZGYA}}\label{sec:fairkmzgya}

Having compared {\it FairKM} against {\it ZGYA} for fairness in a synthetic setting that was favorable to the latter in the previous section, we now do a more direct comparison here. In particular, we consider comparing the {\it FairKM} and {\it ZGYA} instantiations with each sensitive attribute separately, which offers a more level setting. Figure~\ref{fig:adultaw} illustrates the comparison on the AW evaluation measure over the Adult dataset for each $\mathcal{S}$ attribute with {\it ZGYA(S)} and {\it FairKM(S)} values shown separated by the {\it FairKM (All)} value in between them; all these are values obtained with $k=5$. The {\it FairKM (All)} is simply {\it FairKM} instantiated with all attributes in $\mathcal{S}$, which was used in the comparison in the previous section. As may be seen, with {\it FairKM(S)} focusing on just the chosen attribute (as opposed to {\it FairKM (All)} that needs to spread attention across all attributes in $\mathcal{S}$), {\it FairKM(S)} is able to achieve better values for AW. Thus, {\it FairKM(S)} is seen to beat {\it ZYGA(S)} by larger margins than {\it FairKM (All)}, as expected. The {\it Race} attribute shows a different trend, with {\it FairKM(S)} recording a slightly higher AW than {\it FairKM(S)}. While we believe this is likely to be due to an unusually high skew in the {\it race} attribute where $87\%$ of objects take the same single value, this warrants further investigation.  Figure~\ref{fig:adultmw} presents the corresponding chart for MW evaluation measure, and offers similar high-level trends as was observed for AW. The corresponding charts for the Kinematics dataset appear in Figures~\ref{fig:kinematicsaw} and~\ref{fig:kinematicsmw} respectively. Over the much smaller Kinematics dataset, the gains by {\it FairKM(S)} over {\it FairKM (All)} are more pronounced in MW with both techniques recording reasonably similar AW numbers. The observed trends were seen to hold for AE and ME evaluation measures as well, those charts excluded for brevity. To summarize the findings across the datasets, it may be seen that {\it FairKM(S)} may be seen to beat the {\it ZGYA(S)} baseline with larger margins than {\it FairKM (All)} on an average, as desired. 

\begin{figure}[tb]
\centering
\includegraphics[width=\columnwidth]{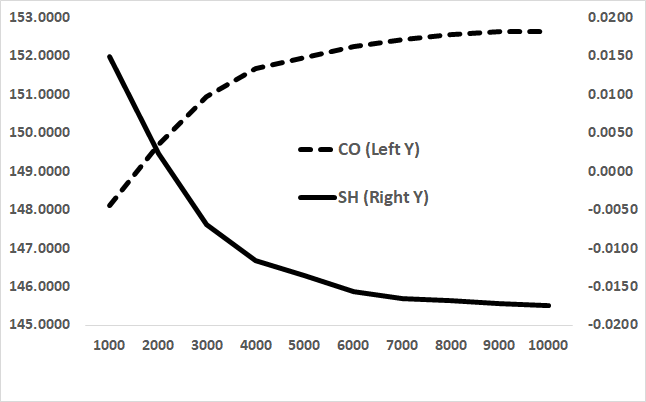}
\caption{Kinematics Dataset: (CO and SH) vs. $\lambda$}
\label{fig:cosh}
\end{figure}

\begin{figure}[tb]
\centering
\includegraphics[width=\columnwidth]{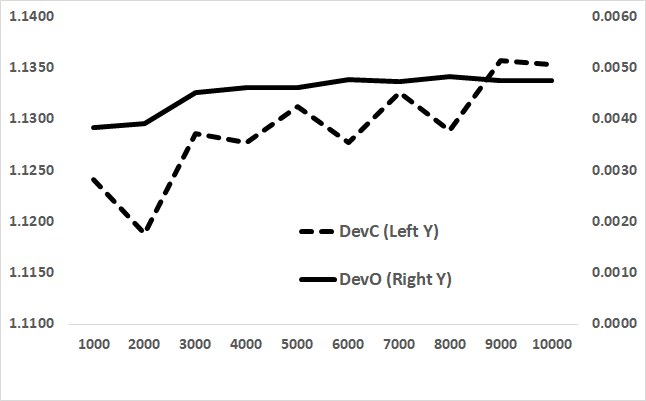}
\caption{Kinematics Dataset: (DevC and DevO) vs. $\lambda$}
\label{fig:devco}
\end{figure}

\begin{figure}[tb]
\centering
\includegraphics[width=\columnwidth]{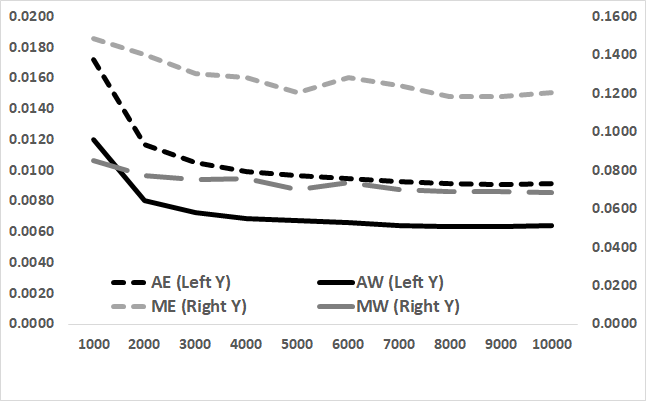}
\caption{Kinematics Dataset: Fairness Metrics vs. $\lambda$}
\label{fig:fair}
\end{figure}

\subsection{{\it FairKM} Sensitivity to $\lambda$}\label{sec:lambdasensitivity}

We now study {\it FairKM}'s sensitivity to it's only parameter $\lambda$, the weight for the fairness term. With increasing $\lambda$, we expect {\it FairKM} to fare better on fairness measures with corresponding degradations in the clustering quality measures. The vice versa is expected to hold with decreasing $\lambda$. We observed such desired trends across Adult and Kinematics datasets, with changes being slower and steadier for the larger Adult dataset. This is on expected lines with the number of parameters such as clustering assignments being larger on the Adult dataset. In the interest of focusing on the smaller dataset, we outline the changes with $\lambda$ on clustering quality and fairness measures on the Kinematics dataset, when $\lambda$ is varied from $1000$ to $10000$. The variations on the CO and SH measures are illustrated in Figure~\ref{fig:cosh}, whereas the variations on DevC and DevO are plotted in Figure~\ref{fig:devco}. We use both sides of the Y-axis to plot the measures which widely vary in terms of their ranges; the axis used is indicated in the legend. As may be seen from them, CO, SH and DevO record slow and steady degradation (the Y-axis is stretched to highlight the region of the change; it may be noted that the quantum of change is very limited) with increasing $\lambda$. The degradation in DevC, however, is more jittery, while the direction of change remains on expected lines. The fairness deviation measures are plotted against varying $\lambda$ (once again, on both Y-axes) in Figure~\ref{fig:fair}. They record gradual but steady improvements (being deviations, they are better when low) with increasing $\lambda$, on expected lines. Overall, it may be seen that {\it FairKM} moves steadily but gradually towards fairness with increasing $\lambda$, as desired.

\section{Conclusions and Future Work}

We considered the problem of ensuring representational fairness in clustering for scenarios comprising multiple sensitive attributes. We proposed a novel clustering method, {\it FairKM}, to accomplish the fair clustering task. In particular, {\it FairKM} comprises the optimization of the classical clustering objective in tandem with a novel fairness loss term, towards achieving a trade-off between clustering quality (on the non-sensitive attributes) and cluster fairness (on the sensitive attributes). We outline a series of evaluation metrics for both the above criteria, and perform a rigorous empirical evaluation of {\it FairKM} over two real-world datasets of widely varying character, pitching {\it FairKM} against other available methods. Our empirical evaluation illustrates that {\it FairKM} outperforms the baseline {\it ZGYA} even over synthetic settings that are artificially favorable to the latter. This illustrates the effectiveness of the {\it FairKM} formulation.

\subsection{Future Work}

We are exploring three directions of future work towards enhancing {\it FairKM}. First, we are studying the performance trends of {\it FairKM} with increasing number of sensitive attributes as well as increasing number of values per sensitive attribute. Second, drawing cue from the observation in Section~\ref{sec:fairkmzgya}, we are looking at how {\it FairKM} can be improved to ensure good performance even on attributes with highly skewed distributions. Third, the main computational bottleneck in {\it FairKM} is the cluster centroid update while doing cluster assignments. We are considering approximation heuristics such as mini-batch updates where centroid updates are done only once every mini-batch of clustering assignment updates, to speed up {\it FairKM} for scalability. 


\noindent{\bf Acknowledgements:} The   second   author   was   partially   supported   by SPARC  (P620).  The  authors  would  also  like  to  thank Krishna M. Sivalingam (IIT Madras) for his constant encouragement.

\bibliographystyle{ACM-Reference-Format}
\bibliography{sample-base}

\end{document}